%% file: main.tex
\documentclass[runningheads]{llncs}

\usepackage[final,year=2026,ID=9481]{eccv}

\usepackage{eccvabbrv}

\makeatletter
\def\@thefnmark{*}
\renewcommand\@makefnmark{\hbox{\textsuperscript{*}}}
\makeatother

\usepackage{graphicx}
\usepackage{booktabs}

\usepackage[accsupp]{axessibility}  

\usepackage{hyperref}

\usepackage{orcidlink}
\usepackage{marvosym}

\usepackage{bbding}

\input{preamble}

\begin{document}

\title{AIMold: An Autonomous AI-based Pipeline for Complex Mold Design}

\titlerunning{AIMold}

\author{Pengyun Qiu\inst{1}\thanks{Equal contribution. \textsuperscript{\Envelope} Corresponding author.}\orcidlink{0009-0009-6090-0258} 
Shuo Wang\inst{1}\textsuperscript{*}\orcidlink{0009-0001-1699-1887} 
Zeyuan Chen\inst{1}\orcidlink{0009-0001-2185-2283}\\ 
Yihao Zhi\inst{1}\orcidlink{0009-0007-1183-5459}
Chongjie Ye\inst{3,1}\orcidlink{0000-0002-7123-0220} 
Xiaoguang Han\inst{1,2,3}\textsuperscript{\Envelope}\orcidlink{0000-0003-0162-3296}
}

\authorrunning{P. Qiu, S. Wang, Z. Chen, Y. Zhi, C. Ye, X. Han}

\institute{School of Science and Engineering, The Chinese University of Hong Kong, Shenzhen  \and Guangdong Provincial Key Laboratory of Future Networks of
Intelligence \and FNii-Shenzhen\\
\email{pengyunqiu0525@yeah.net, shuowang1@link.cuhk.edu.cn, hanxiaoguang@cuhk.edu.cn}\\
\url{https://tb2-sy.github.io/aimold/}}

\maketitle

\begin{figure}[h]
  \centering
  \includegraphics[width=0.8\textwidth]{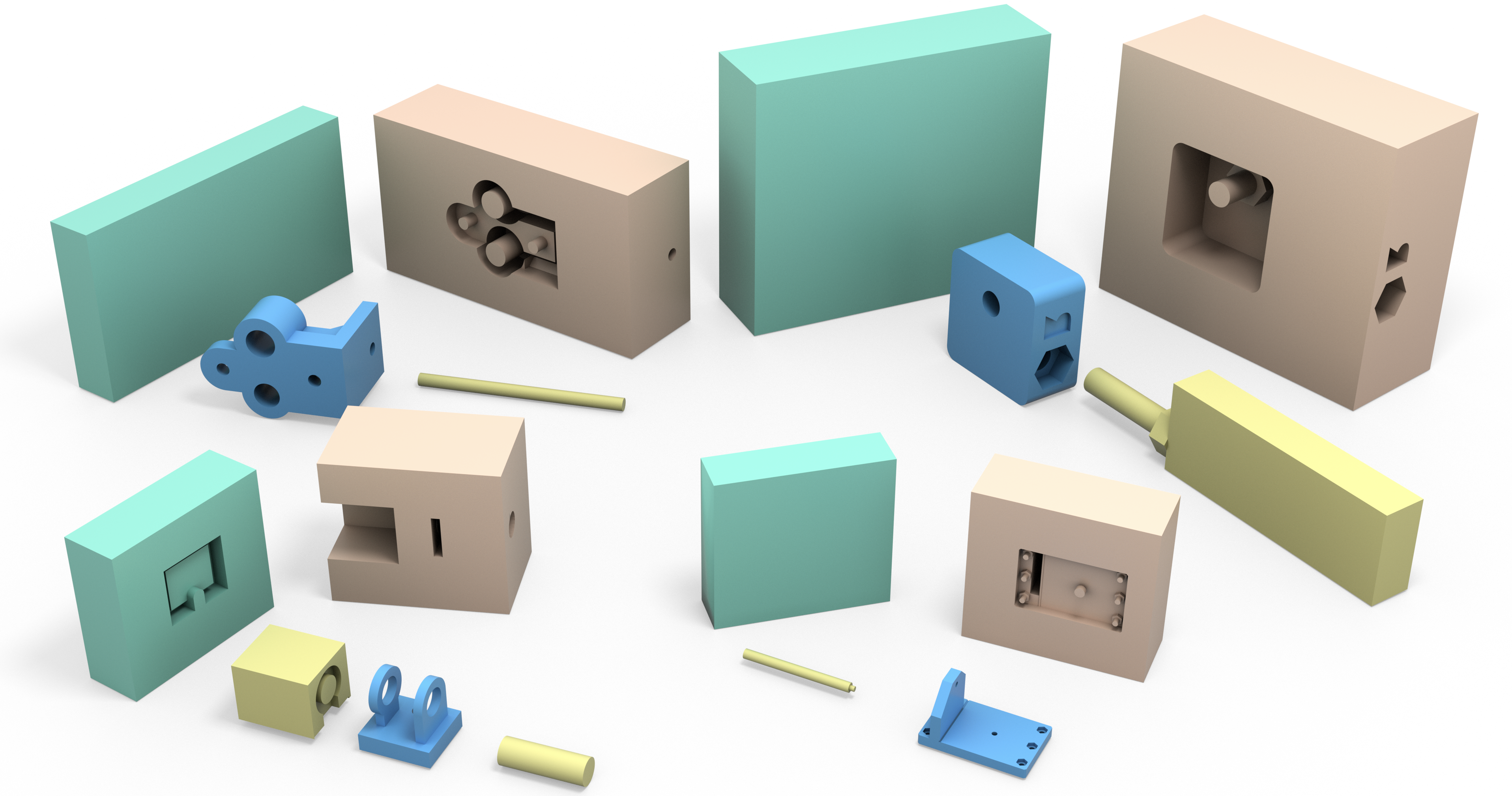}
  \caption{Given a single-body CAD model (blue), AIMold generate high quality assembly mold parts, including auxiliary components (yellow), upper and lower molds (cyan/brown).}
  \label{fig:teaser}
\end{figure}

\vspace{-3.5em}
\begin{abstract}

\input{sec/0_abstract}
\end{abstract}

\input{sec/1_intro}

\input{sec/2_related}
\input{sec/3_method}

\input{sec/4_dataset}
\input{sec/5_experiment}

\input{sec/6_limitation_futurework}
\input{sec/7_conclusion}
\input{sec/8_ack}

\bibliographystyle{splncs04}
\bibliography{references}
\input{sec/X_suppl}
\end{document}

%% file: preamble.tex
\newcommand{\object}{\mathcal{O}}
\newcommand{\numpts}{L}
\newcommand{\voxelencoder}{\boldsymbol{\mathcal{E}}}
\newcommand{\voxeldecoder}{\boldsymbol{\mathcal{D}}}
\newcommand{\voxelfeats}{\boldsymbol{f}}
\newcommand{\latent}{\mathbf{z}}

\newcommand{\latentzero}{\mathbf{z}_{0}}
\newcommand{\noise}{\boldsymbol{\epsilon}}

\newcommand{\timestep}{t}

\providecommand{\Eqref}[1]{Eq.~\eqref{#1}}
\AtBeginDocument{%
  \providecommand{\citet}{\cite}%
  \providecommand{\citep}{\cite}%
}

%% file: sec/0_abstract.tex
Injection molding is the cornerstone of mass-producing plastic components. While current algorithms can automate mold design for basic geometries using standard two-piece molds, complex parts featuring undercuts, side holes, or re-entrant features present a significant challenge. These geometries often necessitate auxiliary components beyond the primary upper and lower molds. In practice, designing these intricate assemblies is a laborious process that relies heavily on expert knowledge. Furthermore, the scarcity of public datasets has hindered the development of effective learning-based solutions. To bridge these gaps, we introduce \textbf{MoldCAD}, a curated dataset that pairs complex single-body CAD parts with industry-standard mold assemblies. Each entry includes the upper and lower molds, parting surfaces, demolding orientations, and necessary auxiliary components. The dataset comprises \textbf{4,934} CAD models and over \textbf{3,850} mold assemblies, totaling more than \textbf{23k} individual models. Building upon this dataset, we propose a comprehensive pipeline that predicts demolding orientations, identifies auxiliary components, and constructs parting surfaces to derive a complete, manufacturing-ready mold assembly for downstream CAD/CAM workflows. Our results demonstrate a promising path toward fully automated industrial mold design and contribute to the broader advancement of manufacturing-aware CAD generation. 

%% file: sec/1_intro.tex
\section{INTRODUCTION}
\begin{figure}[h]
  \centering
  \begin{subfigure}[b]{0.48\columnwidth}
    \centering
    \includegraphics[width=\linewidth]{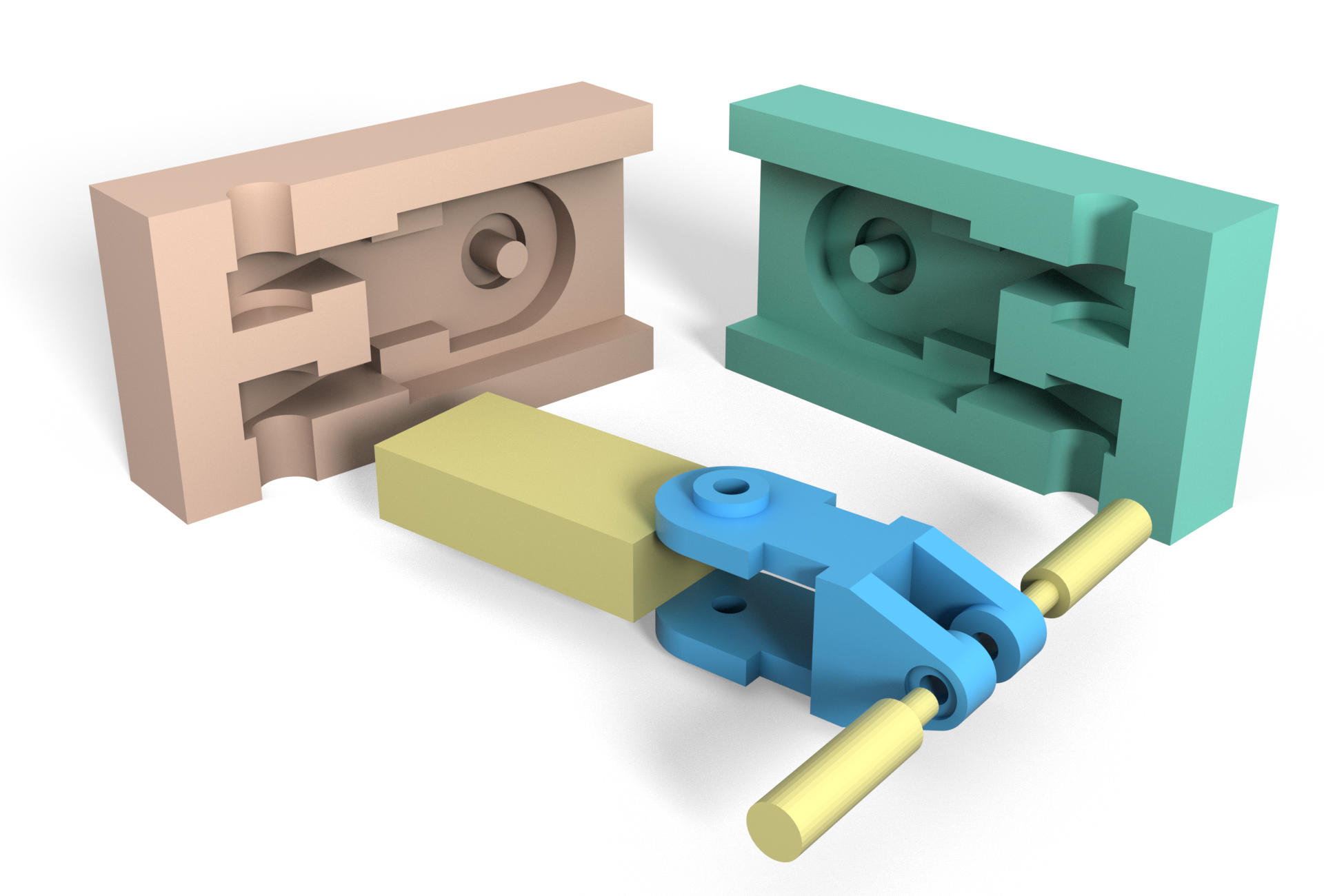}
    \caption{Exploded view} 
    \label{fig:mold_exploded}
  \end{subfigure}
  \hfill 
  \begin{subfigure}[b]{0.48\columnwidth}
    \centering
    \includegraphics[width=\linewidth]{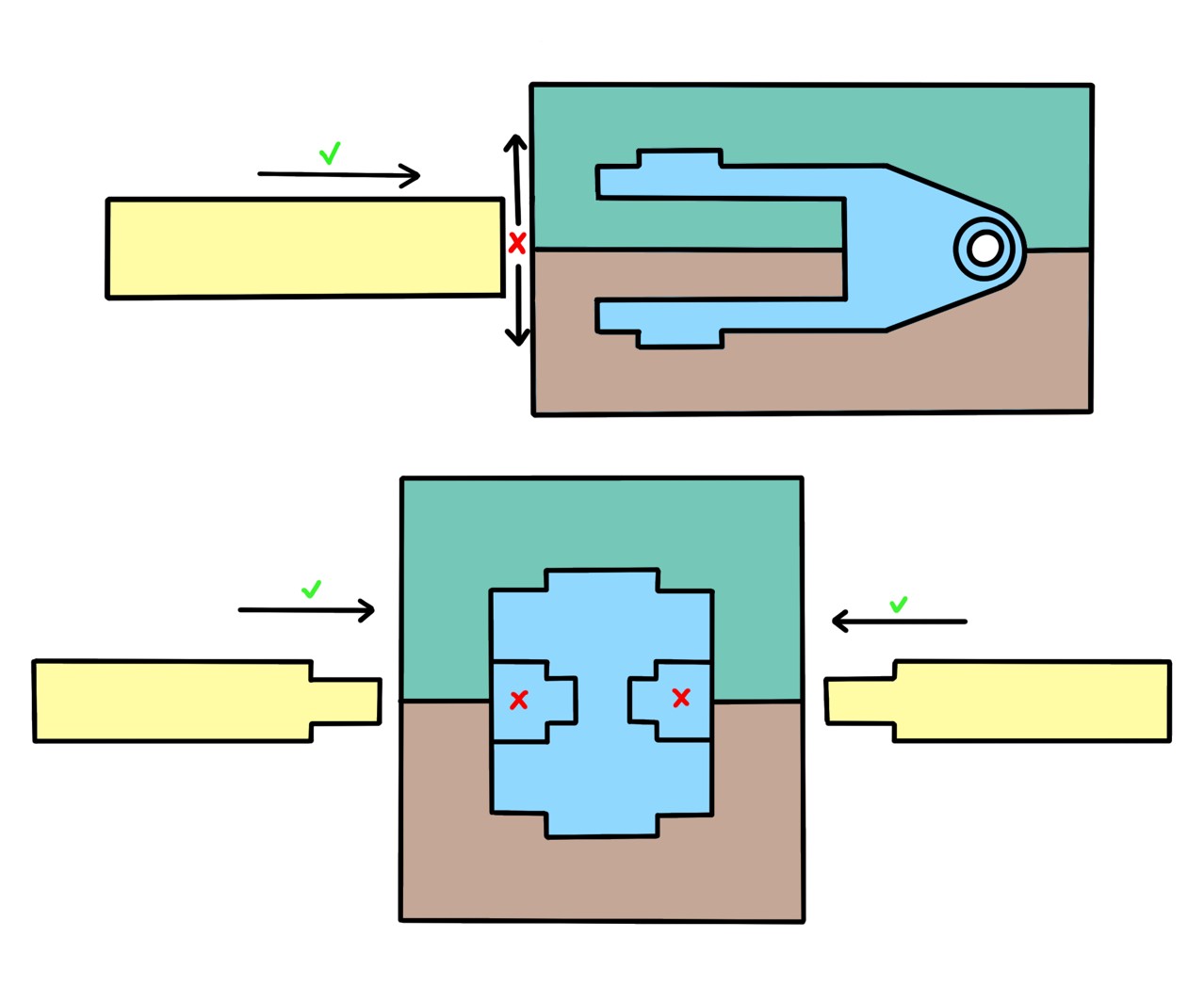}
    \caption{Auxiliary components}
    \label{fig:side_view}
  \end{subfigure}
  
  \caption{Mold assembly details. (a) Exploded view showing the input part (blue), molds (cyan/brown), and auxiliary components (yellow). (b) Geometric necessity: Auxiliary components provide lateral freedom for collision-free demolding.}
  \label{fig:mold_combined}
\end{figure}

Injection molding is the foundation of global industrial manufacturing. It facilitates the production of a vast array of products, from small components like keyboard keycaps and bottle caps to large-scale items such as integrated chairs and fan housings. In this process, heated materials, like resins or thermoplastics, are injected into a precision mold, solidifying into a specific shape. A well-made mold can typically endure millions of production cycles. However, while the physical fabrication stage is largely automated, mold design persists as a significant bottleneck in many computer-aided manufacturing (CAM) workflows.

In traditional industrial practice, mold design is a staged workflow executed by experienced engineers. The process begins with selecting an optimal demolding orientation, followed by the layout of a parting surface to bifurcate the upper and lower molds. Finally, engineers must integrate auxiliary components whenever undercuts or side holes obstruct the demolding path of a standard two-piece mold (Figure~\ref{fig:side_view}). The reliance on domain expertise creates a significant bottleneck, limiting the scalability and efficiency of the broader manufacturing pipeline.

Automating this assembly generation remains challenging. We focus on a prevalent setting as shown in Figure~\ref{fig:mold_exploded}. Given a single-body part as input, our objective is to generate a comprehensive assembly comprising the upper and lower molds, the parting surface, and a variable number of auxiliary components. This scope specifically addresses the scenarios that dominate manual design: parts with undercuts and side holes. Such features are usually incompatible with two-piece molds and require the incorporation of auxiliary components.


Data-driven approaches, particularly deep learning, hold significant promise for capturing complex design regularities. However, prior research has typically targeted isolated subproblems—such as determining demolding directions ~\cite{chakraborty2009automatic, khardekar2005finding} or detecting undercuts ~\cite{ran2010design, banerjee2007geometrical}, rather than addressing the holistic construction of the complete mold assembly. Consequently, these methods often rely on geometric heuristics that are fragile and limited to simple, idealized shapes. As geometric complexity increases, the space of feasible mechanisms expands exponentially, rendering traditional design logic difficult to formalize. In contrast, recent generative models offer a more compelling path forward, as they can learn implicit design patterns that defy formalization and integrate these decisions into a unified generation process. Nevertheless, progress in this direction is fundamentally obstructed by a critical data gap: there are virtually no public datasets that pair single-body parts with professionally designed mold assemblies. To the best of our knowledge, the scarcity of manufacturing-ready CAD datasets containing such paired assemblies makes it nearly impossible to train, evaluate, or even frame mold assembly generation as a viable learning problem. Bridging this gap is essential to achieving a transformative leap in automated mold design \cite{DBLP:journals/cgf/AlderighiMABCP22}.

To address the scarcity of high-quality annotated CAD data, we present \textbf{MoldCAD}, a large-scale paired dataset comprising approximately 5,000 mold assembly examples. Constructing this benchmark presents unique challenges beyond simple data aggregation. Specifically, we manually selected candidate parts that satisfy strict manufacturing prerequisites—ensuring they are manifold, watertight, and solid—while specifically retaining complex features such as undercuts and side holes. To ensure diversity, we curated the collection to eliminate near-duplicate geometries. For every part, we provide a complete mold assembly authored by professional engineers, accompanied by consistent semantic annotations for every component within the assembly. This dataset enables researchers to learn and benchmark the generation of both the final CAD shapes and the underlying mechanical structures required for successful demolding.

Building upon MoldCAD, we formulate mold assembly generation as a conditional generation problem, which includes auxiliary components, upper and lower mold generation. Given an input CAD model, we need to estimate its optimal orientation for subsequent manufacturing processes. We employ a plane-detection-based method to generate a candidate set of orientations and train a binary classifier using manually curated positive and negative samples to select the most suitable orientation. For assembly generation, we devise a two-stage framework. In the first stage, a coarse structural representation is generated in the voxel domain to establish the global layout and component relationships. In the second stage, we refine this coarse draft into detailed assembly meshes, recovering the fine geometric detail that the low-resolution first stage cannot capture. These meshes are then regularized based on the input model through geometry processing routines and exported as CAD-friendly representations for downstream CAD/CAM verification, without introducing additional learned components.

In summary, we make the following contributions:
\begin{itemize}
    \item We raise the challenge of mold manufacturing design and, for the first time, employ deep learning to develop a comprehensive, step-by-step approach. This approach covers demolding orientation estimation, auxiliary component generation, upper and lower mold generation, and the conversion of generated meshes into CAD models.
    \item To support this approach, we introduce \textbf{MoldCAD}, a carefully curated dataset designed specifically for the mold design task, which includes rich annotations such as auxiliary components, parting surfaces, demolding orientations, and paired upper and lower molds.
    
    \item Extensive experiments demonstrate the promising results of our data-driven approach to transform complex mold design into an automated process, substantially reducing costs associated with traditional industrial design workflows.
\end{itemize}

%% file: sec/2_related.tex
\section{RELATED WORK}
\subsection{Mold design}
Recent advancements in digital fabrication technology have driven research on mold design in the fields of computer-aided design, computer graphics, and mechanical engineering.
Early methods addressed moldability by optimizing accessibility metrics, including undercut area, parting-surface flatness, and parting depth ~\cite{chakraborty2009automatic, khardekar2005finding}. Later studies formalized these concepts through visibility analysis, leveraging volumetric or mesh-based discretizations to produce dense demoldability maps and valid manufacturing annotations ~\cite{mercado2016new, donate2013new, fu2008application}.


Rigid molds require strict geometric compliance, motivating extensive work on automating the search for feasible parting directions and lines/surfaces. \cite{DBLP:journals/cgf/HerholzMA15} approximate freeform shapes to minimize the number of mold pieces, and~\cite{DBLP:journals/cg/SteinJG19} deform inputs into shapes castable with two-piece rigid molds, while~\cite{DBLP:journals/tog/AlderighiMBCP21} segment complex objects into height-field parts via volumetric decomposition without altering the geometry.

For models that violate two-piece constraints, multi-piece molds and side-actions map undercut features to candidate retraction spaces, often via graph-based search or cost optimization~\cite{ran2010design, banerjee2007geometrical, fu2008application}. More complex parts are partitioned by reasoning about core regions and generating parting curves through geometry- and graph-based analyses~\cite{hou2018hybrid, lin2014automatic}.

Parallel efforts use flexible materials and composite molds: \citet{DBLP:journals/tog/MalomoPBC16} generate peelable flexible shells via dynamic simulation, and \citet{DBLP:journals/tog/AlderighiMGPBC18} compute optimal cut layouts for silicone molds. As surface-based analysis often fails on intricate geometries, \citet{DBLP:journals/tog/AlderighiMGBCP19} combine a rigid outer shell with a soft inner part defined by volumetric escape-path analysis. Beyond casting, \citet{DBLP:journals/tog/ZhangFSGWW19} optimize flat-panel shapes and supports to cast plaster sculptures via fluid-pressure deformation.

\subsection{Learning-based CAD modeling}
Our work also builds on learning-based CAD generation, which has evolved from recovering modeling sequences to directly generating boundary representations (B-Reps). Early methods treated CAD modeling as sequence prediction, recovering the construction history (e.g., sketch-and-extrude operations): Transformer-based models generate command sequences~\cite{wu2021deepcad}, later refined with decoupled or hierarchical codebooks for finer geometric and topological control~\cite{xu2022skexgen,xu2023hierarchical}.

In particular, ~\cite{you2025img2cad} parse discrete CAD commands through Visual Language Models (VLMs). Similarly, sketch-based approaches either parse hand-drawn strokes into parametric CAD command sequences~\cite{li2020sketch2cad,li2022free2cad,xu2024cad,qi2026pointer} or directly generate constraint-aware sketches~\cite{para2021sketchgen,seff2021vitruvion}.
Although sequence-based approaches ensure validity, they face a fundamental one-to-many mapping challenge: the same final geometry may admit many different operation sequences, which complicates learning and evaluation. A parallel stream of research investigates the direct generation of B-Reps. ~\cite{xu2025autobrep} and ~\cite{jayaraman2022solidgen} utilize autoregressive Transformers and a unified tokenization scheme to achieve end-to-end generation of vertices, edges, and faces. To treat topology and geometry independently, diffusion-based and latent variable models~\cite{xu2024brepgen,lee2025brepdiff,guo2025brepgiff,liu2025hola,li2025stitch} have been widely used.

Other methods reconstruct editable CAD directly from raw geometry. Self-supervised approaches recover shapes by assembling primitives~\cite{yu2022capri} or by learning sketch, feature, and machining operations~\cite{li2023secad,li2024sfmcad,yavartanoo2024cnc}, whereas reverse engineering pipelines fit parametric models to point clouds~\cite{liu2024point2cad,khan2024cad,dupont2024transcad}.

Multimodal conditioning further broadens usability: language and image conditioned generation builds structured representations through LLMs and multimodal training~\cite{zhang2024flexcad,alam2024gencad,khan2024text2cad,xu2024cad}, and text-guided B-Rep editing operates directly in a learned latent space without the construction history~\cite{liu2025b}. For representation learning, \citet{jayaraman2021uv} proposes a grid-graph architecture for B-Rep classification and segmentation, while \citet{ma2023multicad} aligns sequence and geometry representations via contrastive learning. Beyond single parts, \cite{willis2022joinable} mates parts into constrained assemblies.

\subsection{3D generative models}
Early 3D generation methods primarily relied on Generative Adversarial Networks (GANs)~\cite{chan2022efficient,gao2022get3d,skorokhodov20233d,wu2016learning} for 3D modeling. Although these approaches demonstrated the feasibility of learning 3D representations~\cite{deng2021gram,zheng2022sdf}, they often suffered from limited generation quality and training instability.
The emergence of diffusion models~\cite{ho2020denoising,sohl2015deep,nichol2022point,muller2023diffrf} marked a conceptual shift in 3D generation, leading to a substantial improvement in generation quality.
In the field of image generation~\cite{zhang2023adding,rombach2022high,lin2023magic3d}, diffusion models have achieved remarkable success. This progress has sparked strong interest in lifting 2D generative models to 3D, leading to the emergence of Score Distillation Sampling (SDS)~\cite{poole2023dreamfusion,liang2024luciddreamer,tang2023make,tang2024dreamgaussian}. However, SDS suffers from substantial additional optimization time and limited multi-view consistency. Large Reconstruction Models~\cite{hong2024lrm,li2024instant3d,liu2024one,liu2024meshformer} address the per-scene optimization issue by enabling generation through a single forward pass. With the continued expansion of large-scale 3D datasets~\cite{deitke2023objaverse,yu2023mvimgnet}, improving the scalability of 3D latent diffusion models has become a central research focus, and many representative works~\cite{zhang20233dshape2vecset,zhang2024clay,zhao2023michelangelo,trellis,xiang2025native,zhao2025hunyuan3d,cube3d,lai2025lattice} have achieved significant progress in efficiently generating high-quality 3D shapes and structures. Different from these large-scale 3D generation methods, such performance is currently difficult to achieve in the mold design domain due to the limited size of available datasets. Therefore, our proposed method, AIMold, explores 3D generation under dataset constraints and achieves promising results.

%% file: sec/3_method.tex
\section{METHOD}
\subsection{Problem definition}

We formulate injection mold design as a conditional generation task that maps a single-body part to an assembly that can reproduce the part geometry and support demolding. We focus on complex parts with undercuts and side holes, where a two-piece mold is often insufficient and auxiliary components are required in addition to the upper and lower molds.

\paragraph{Input}Let $P$ denote a single-body part represented as a CAD solid or a watertight mesh. Our pipeline operates on a voxelized representation $V(P)\in\{0,1\}^{N \times N \times N}$, obtained by rasterizing $P$ in a canonical coordinate frame.
\paragraph{Output}The target output is an assembly
\[
  \mathcal{A} = \{ M_u, M_l, S_{part}, \mathcal{C} \}
\]
where $M_u$ and $M_l$ are the upper and lower molds, $S_{part}$ is the parting surface, and $\mathcal{C} = \{C_1, \dots, C_K\}$ are the auxiliary components. Note that $K$ is not fixed but implicitly determined by the geometric features of $P$. Each element in $\mathcal{A}$ is represented as a high quality mesh to support downstream operations.

\paragraph{Challenges}This task is challenging for three reasons. First, the output is structured: the assembly consists of multiple interacting elements rather than an isolated entity. Second, the mapping from $P$ to $\mathcal{A}$ is inherently one-to-many. Different demolding directions, parting surfaces, and auxiliary-component configurations can all be acceptable for the same part, which makes direct regression ill-posed. Third, paired data is scarce and highly niche: industrial mold assemblies follow strong design regularities that are difficult to design manually without expert knowledge.

Accordingly, we learn a conditional generative model $p(\mathcal{A}\mid V(P))$ from engineering designs. Our goal is to generate geometrically plausible assemblies that match common industrial patterns and can serve as structured drafts for downstream CAD/CAM verification and refinement. We do not explicitly enforce demolding constraints during generation; instead, we rely on the implicit manufacturability priors present in the training data.

\begin{figure*}[t]
  \centering
  \includegraphics[width=1.0\textwidth]{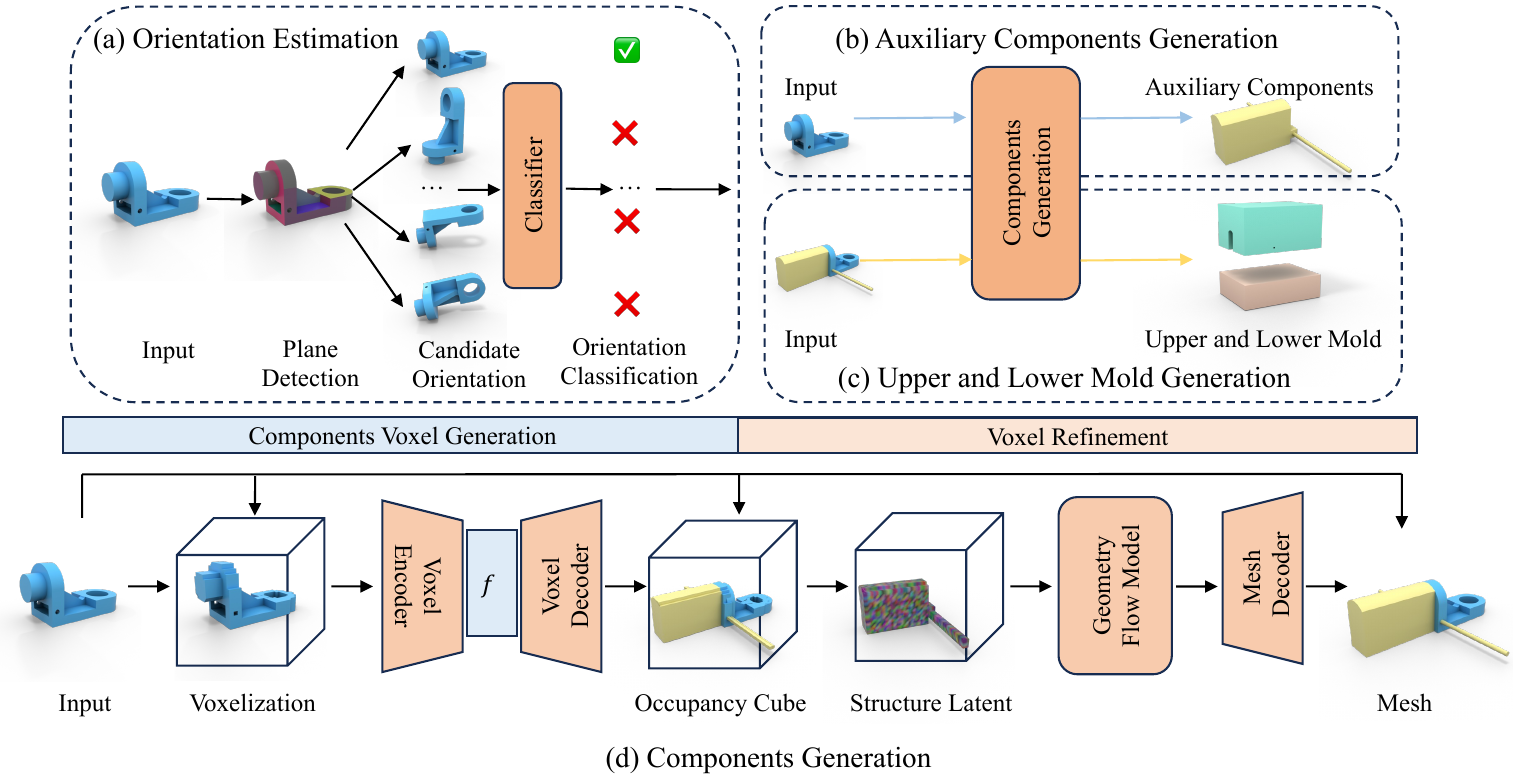}
\caption{Pipeline overview of our method. The framework first estimates the optimal model orientation via plane detection and classification (a), followed by the generation of auxiliary components (b), and upper and lower molds (c). The detailed generation process (d) operates in two sequential phases: Components Voxel Generation, which predicts a coarse occupancy cube from the voxelized input, and Voxel Refinement, which leverages a geometry flow model on structure latents to synthesize the faithful mesh.}
  \label{fig:pipeline_single}
\end{figure*}

\subsection{Components Generation}
\label{sec:3_2}
We propose a two-stage pipeline for generating auxiliary components, upper and lower molds. Figure~\ref{fig:pipeline_single} shows an overview, with details described below. We first use a voxel encoder $\voxelencoder$ to encode voxelized 3D model $\object$ to features $\voxelfeats$ and decode features using a voxel decoder $\voxeldecoder$ as shown in Figure~\ref{fig:pipeline_single} (d). The voxel decoder's outputs will be refined through our voxel refinement stage. This stage predicts the 3D shape in the occupied regions based on the outputs from the first stage. In the second stage, we generate latents $\{\boldsymbol{z}_i\}_{i=1}^{L}$ given the structure $\{\boldsymbol{p}_i\}_{i=1}^{L}$ using a transformer $\boldsymbol{\mathcal{G}}_{\mathrm{L}}$ designed for sparse structures, where $\boldsymbol{p}_i \in \mathbb{R}^3$ denotes the spatial coordinates of the $i$-th active voxel within a voxel grid, and $\boldsymbol{z}_i$ encodes the associated voxel features. We use Rectified Flow~\cite{lipman2022flow} for the latent $\latent$ generation. 
\vspace{-1pt}
\begin{equation}
    \mathcal{L}_{FM}(\theta)=\mathbb{E}_{t,\boldsymbol{z}_0,\boldsymbol{\epsilon}}\|\boldsymbol{v}_\theta(\boldsymbol{z}_t, t)-(\boldsymbol{\epsilon}-\boldsymbol{z}_0)\|^2_2. \label{eq:cfm}
\end{equation}
The model interpolates between data $\latentzero$ and noise $\noise$ over time $\timestep$ via $\boldsymbol{z}_t = (1 - \timestep) \boldsymbol{z}_0 + \timestep \noise$. We parameterize $\boldsymbol{v}$ with a neural network $\boldsymbol{v}_{\theta}$ and train it using the Flow Matching (FM) objective as \Eqref{eq:cfm}.

\subsubsection{Representation of 3D Object}
To facilitate network training, we represent each 3D object using a voxel-based formulation with two attributes:
$\object \triangleq \{\boldsymbol{z}_i, \boldsymbol{p}_i\}_{i=1}^{\numpts}$. These features $\{\boldsymbol{z}_i\}_{i=1}^{L}$ are obtained by aggregating multi-view DINOv2 features~\citep{oquab2023dinov2} and subsequently compressed using a 3D variational autoencoder (VAE) encoder. Here, $\numpts$ denotes the total number of active voxels. In the first stage, we rely solely on the voxel coordinates $\boldsymbol{p}_i$ to voxelize the 3D objects. In the second stage, both voxel coordinates $\boldsymbol{p}_i$ and their associated features $\boldsymbol{z}_i$ are used to construct structural latent representations. The structural latents can be decoded into mesh representations~\cite{trellis}.

\subsubsection{Auxiliary Components Generation}
In the components generation stage, a 3D model without auxiliary components is translated into a voxel representation with auxiliary components with our components voxel generation model. Based on a voxelized representation $V(P)\in\{0,1\}^{N \times N \times N}$, coarse geometry with union auxiliary components $\mathcal{C} = \{C_1, \dots, C_K\}$ is produced using a voxel-to-voxel paradigm. Although a CAD model may contain multiple auxiliary components, these components rarely intersect with each other. Therefore, naively predicting multiple auxiliary components within a shared coordinate system is feasible, and they can be separated in a straightforward manner. The voxel encoder and voxel decoder share the same transformer structure like a 3DVAE~\cite{trellis}. In our experiments, we observe that, at the current scale of the Mold dataset, this paradigm aligns more closely with the input 3D models than implicit representation methods~\cite{cube3d,zhao2025hunyuan3d}, as shown in Table~\ref{tab:aux generation_fidelity}.

  

\subsubsection{Upper and Lower Mold Generation}
For CAD models equipped with auxiliary components, we subsequently generate the corresponding upper and lower molds. The upper and lower molds generation procedure follows a pipeline similar to that used for auxiliary component generation, differing only in the input and the generation targets. For auxiliary component generation, we take only $V(P)$ as input and predict the union voxel representation of $C$. In contrast, for upper and lower mold generation, the input consists of the joint voxelized representation of $V(P)$ and $\mathcal{C}$ in a shared voxel coordinate system, and the outputs are the voxel representations of the upper mold $M_u$ and the lower mold $M_l$. Unlike auxiliary components generation, which formulates the output space as a binary occupancy prediction (occupied or empty), upper and lower mold generation requires partitioning the space into two non-overlapping yet fully occupied volumetric results. To this end, we redefine the output space to include three categories: $\text{occupy}_{\text{upper}}$, $\text{occupy}_{\text{lower}}$, and empty.
With this formulation, the upper and lower molds can be generated independently while ensuring spatial consistency.

\subsubsection{Voxel Refinement}
Although direct operations in voxel space can produce reasonable coarse geometry, they are limited by low resolution and lack fine details, making post-processing more challenging.
We employ a geometry flow model to refine the generation of auxiliary components, upper and lower molds. Building upon the approach in \cite{trellis}, we adapt the structural latent generation stage to refine voxel structures. By removing the original text/image conditions, we input only noisy voxel representations to generate structured latents $\boldsymbol{z}$. After training on our dataset, geometric features such as curves, planes, and edges are refined to more closely resemble CAD representations. The structured latents will decode into meshes $\mathcal{C}'_\mathcal{M}$ via decoders $\boldsymbol{\mathcal{D}_\mathcal{M}}$~\cite{trellis}. The decoding process is as follows:
\begin{equation}
    \boldsymbol{\mathcal{D}_\mathcal{M}}\!:\{(\boldsymbol{z}_i,\boldsymbol{p}_i)\}_{i=1}^{L}\!\rightarrow\!\{\{(\boldsymbol{w}_i^j, d_i^j)\}_{j=1}^{64}\}_{i=1}^{L},
\end{equation}
where $\boldsymbol{w}_i^j\in\mathbb{R}^{45}$ are the flexible parameters in FlexiCubes~\cite{shen2023flexicubes} and $d_i^j\in\mathbb{R}^{8}$ is signed distance values for the eight vertices of the corresponding voxel.
\subsubsection{Mesh to CAD}
Although our generative model is able to predict the approximate structure of the auxiliary components and molds, the raw output often lacks precision. We introduce a post-processing stage to regularize these geometries based on the contact constraints imposed by the input $P$.

We first convert the raw prediction $\mathcal{C}'_\mathcal{M}$ into a watertight manifold $\widetilde{C}'_\mathcal{M}$ mesh using Poisson surface reconstruction (screened Poisson) ~\cite{kazhdan2013screened}. To identify the orientation, we analyze the local neighborhood of $P$ adjacent to $\widetilde{C}'_\mathcal{M}$ to estimate the dominant axis $\mathbf{d}$. This allows us to robustly segment $\widetilde{C}'_\mathcal{M}$ into an embedded portion and an external portion. For the embedded portion, we extract an intersection polyline or curve from the target cavity and then generate a swept “mating volume” along $\mathbf{d}$. For the external portion, we improve regularity by fitting standard geometry primitives via RANSAC ~\cite{schnabel2007efficient}. The final shape is obtained by merging these segments via boolean composition. Simultaneously, we reconstruct the upper and lower mold ($M_u, M_l$) from the coarse outputs. We estimate a separating parting surface $S_{part}$ (via planar or quadratic fitting) from the interface, then split an inflated bounding volume of the assembly along $S_{part}$. The final molds are carved by the part $P$ and optimized auxiliary components $\widetilde{C}'_\mathcal{M}$ from the respective blocks. Finally all reconstructed models ($M_u,M_l$ and $\widetilde{C}'_\mathcal{M}$) are optimized via ShapeUp~\cite{bouaziz2012shape}, followed by quad-remeshing~\cite{jakob2015instant}. The resulting meshes are converted to NURBS surface patches~\cite{catmull1998recursively} and exported as a STEP/STP file that supports operations and editing in standard CAD software.

\subsection{Orientation Estimation}
Given an input CAD model, we need to estimate its optimal orientation for subsequent manufacturing processes. To this end, we employ a plane-detection-based method to generate a candidate set of orientations, $\mathcal{P}l = \{Pl_1, \dots, Pl_m\}$, effectively discretizing the continuous pose space into a finite set of feasible solutions. Here, $m$ denotes the total number of detected planar regions. Based on these candidate orientations, we train a binary classifier using manually curated positive and negative samples to select the most suitable orientation, as shown in Figure~\ref{fig:pipeline_single}. Specifically, the ground-truth orientation is used to define positive samples: candidate orientations whose rotation axes are aligned with the ground-truth orientation are labeled as positive. In contrast, negative samples are constructed from plane-detection--derived candidate orientations whose rotation axes are not aligned with the ground-truth orientation. To address class imbalance, we apply data augmentation to both positive and negative samples, expanding each class to four times the scale of the original dataset. For the classifier architecture, we adopt the same voxel representation \( V(P) \) and voxel encoder described in Section~\ref{sec:3_2}. The original decoder is replaced with a lightweight binary classifier, which functions similarly to an anomaly detector by distinguishing valid orientations from invalid ones.

%% file: sec/4_dataset.tex
\section{DATASET}
MoldCAD is a paired dataset of single-body parts and professionally designed injection-mold assemblies. Each sample links an input part to an output assembly that includes the upper and lower molds, a parting surface, and a variable number of auxiliary components required for demolding.

\begin{figure*}[t]
    \centering
    
    \begin{minipage}[b]{0.19\linewidth}
        \centering
        \includegraphics[width=\linewidth]{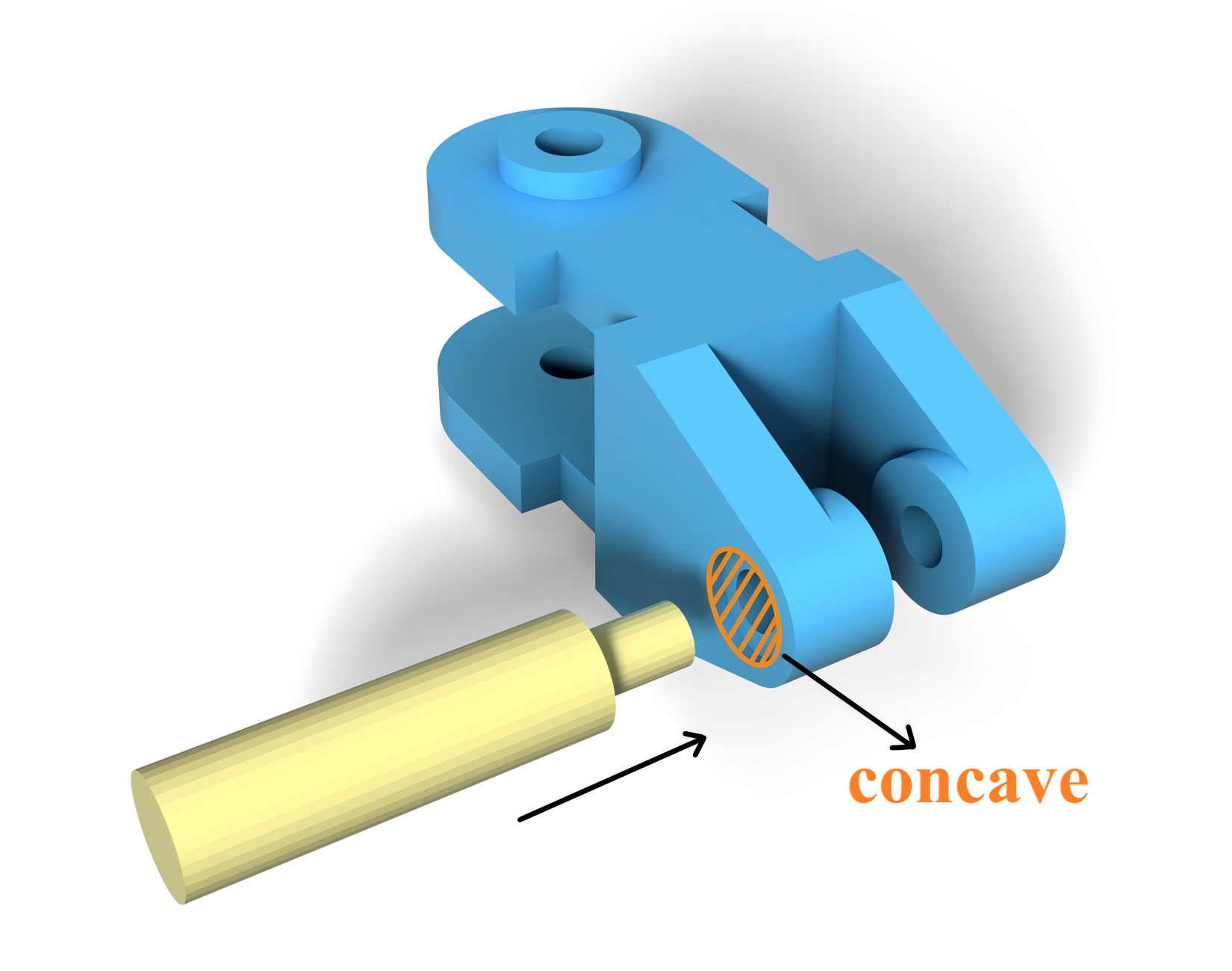}
        \par\vspace{3pt} 
        \small (a) 
    \end{minipage}
    \hfill
    \begin{minipage}[b]{0.19\linewidth}
        \centering
        \includegraphics[width=\linewidth]{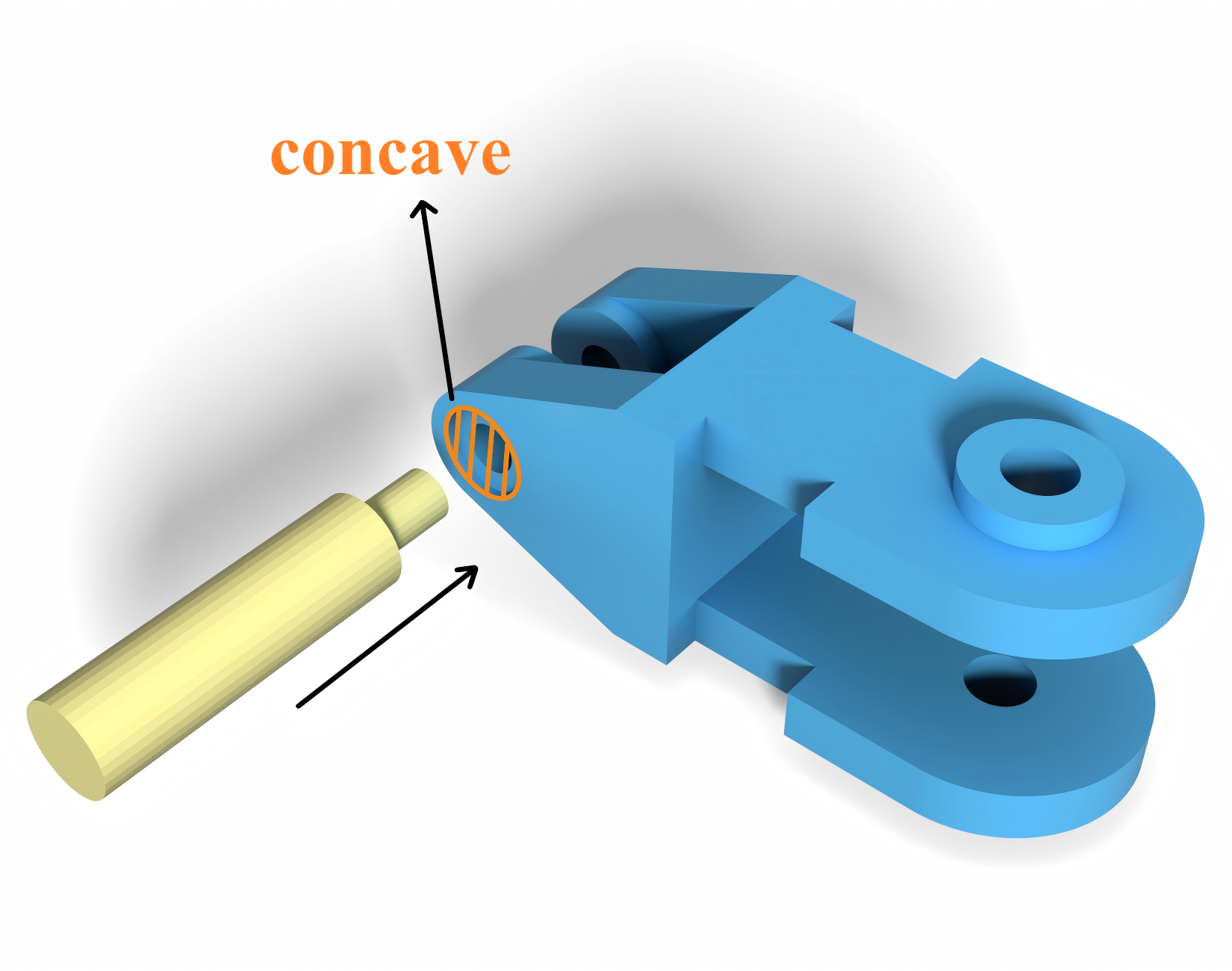}
        \par\vspace{3pt}
        \small (b) 
    \end{minipage}
    \hfill
    \begin{minipage}[b]{0.19\linewidth}
        \centering
        \includegraphics[width=\linewidth]{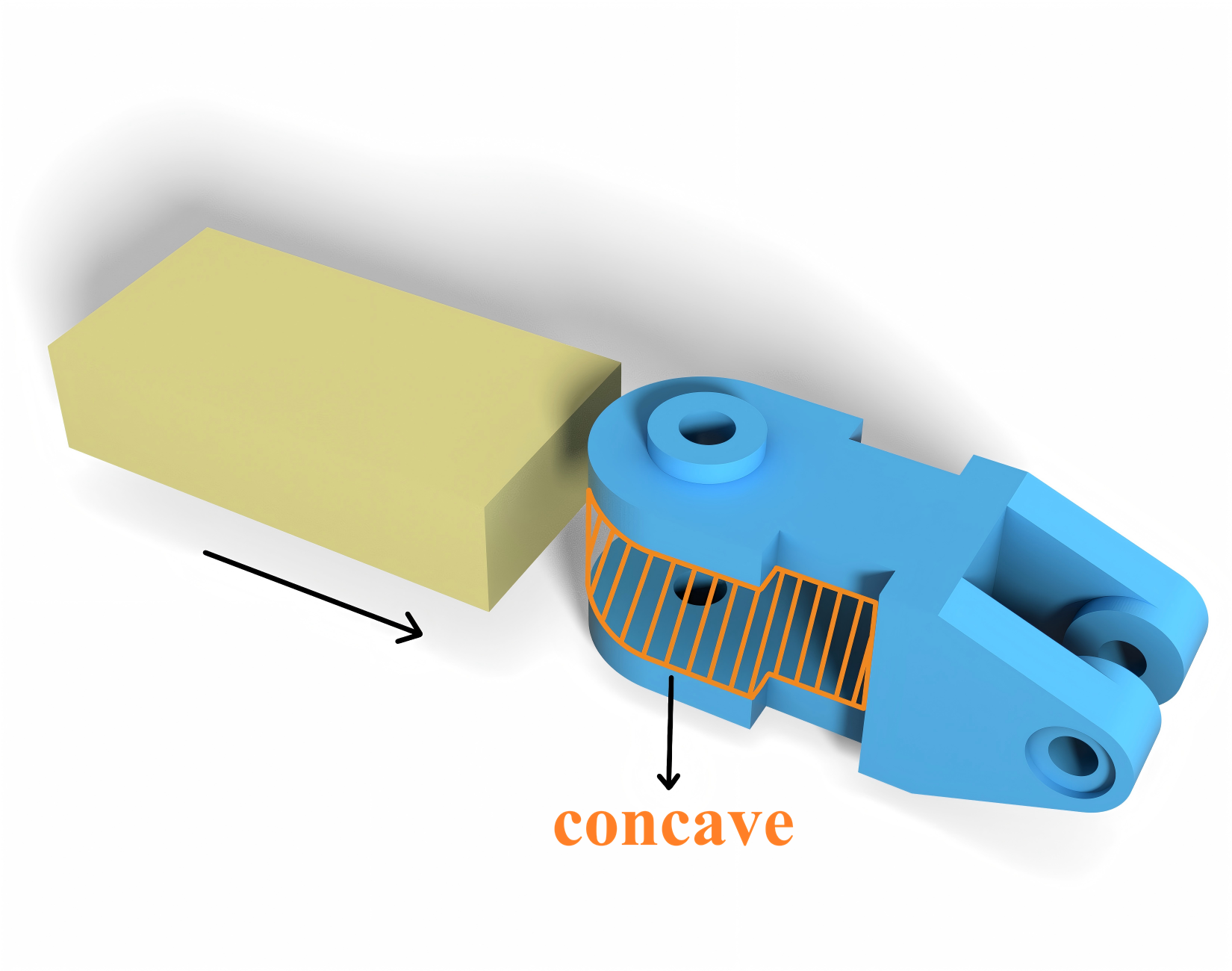}
        \par\vspace{3pt}
        \small (c) 
    \end{minipage}
    \hfill
    \begin{minipage}[b]{0.19\linewidth}
        \centering
        \includegraphics[width=\linewidth]{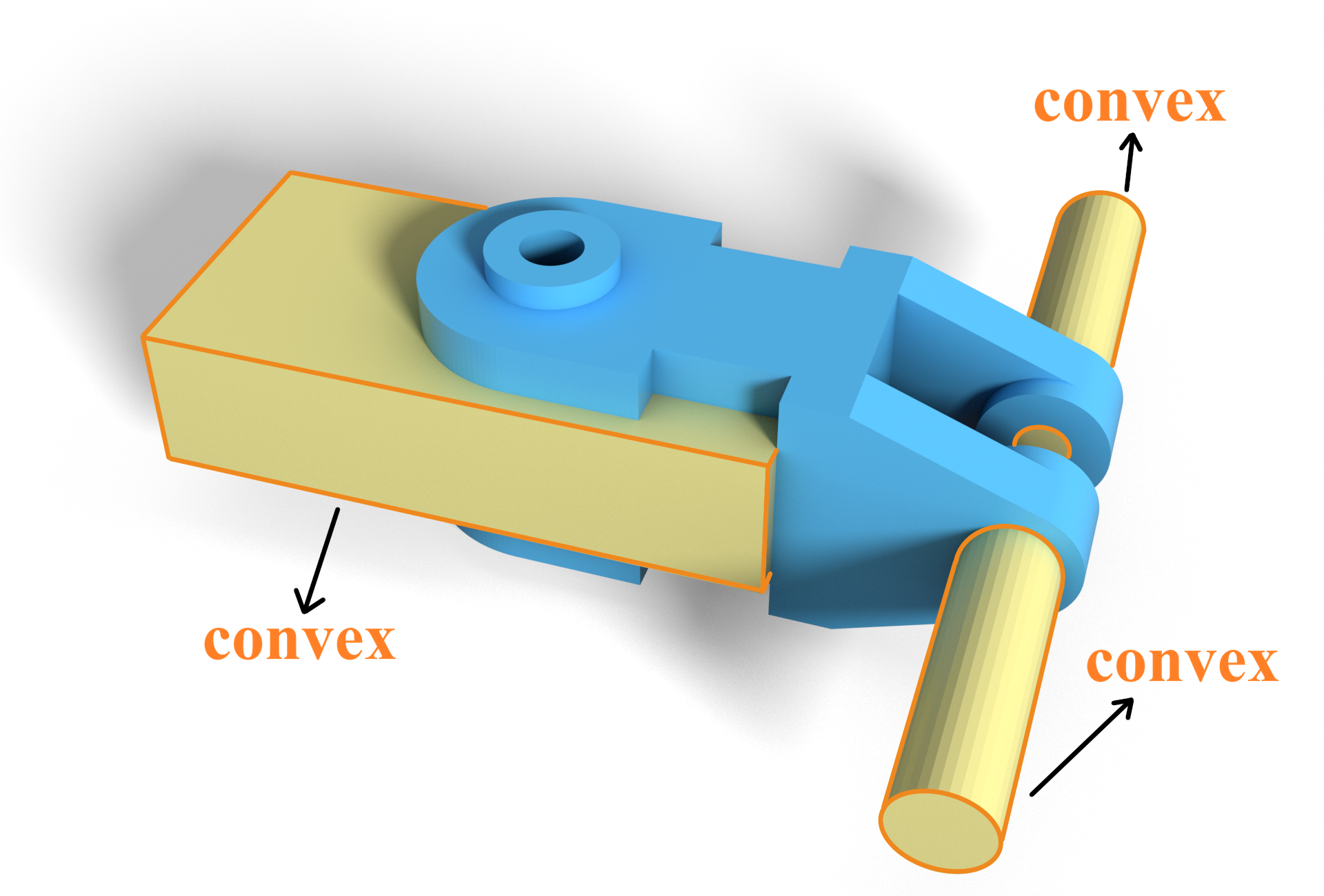}
        \par\vspace{3pt}
        \small (d) 
    \end{minipage}
    \hfill
    \begin{minipage}[b]{0.19\linewidth}
        \centering
        \includegraphics[width=\linewidth]{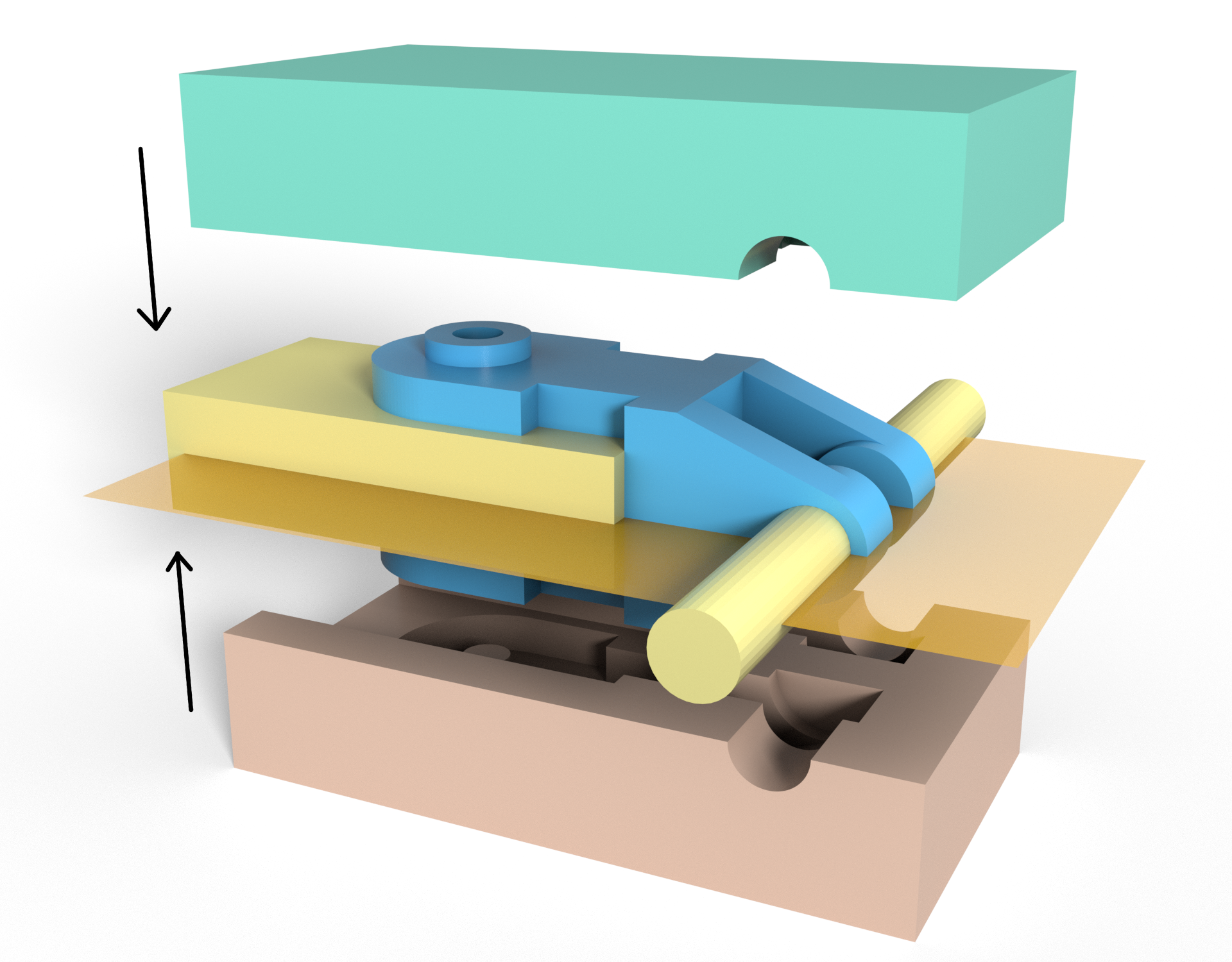}
        \par\vspace{3pt}
        \small (e) 
    \end{minipage}

    \caption{Illustration of the processing pipeline for manual mold assembly design. (a)-(c) Analyze the part along a chosen demolding direction, localize concave (non-demoldable) regions and determine feasible auxiliary components. (d) Generate auxiliary component solids to address these localized collision issues. (e) Create a parting surface to separate upper and lower molds.}
    \label{fig:mold_design_pipeline}
\end{figure*}

\begin{figure}[!t]
  \centering
\includegraphics[width=0.95\columnwidth]{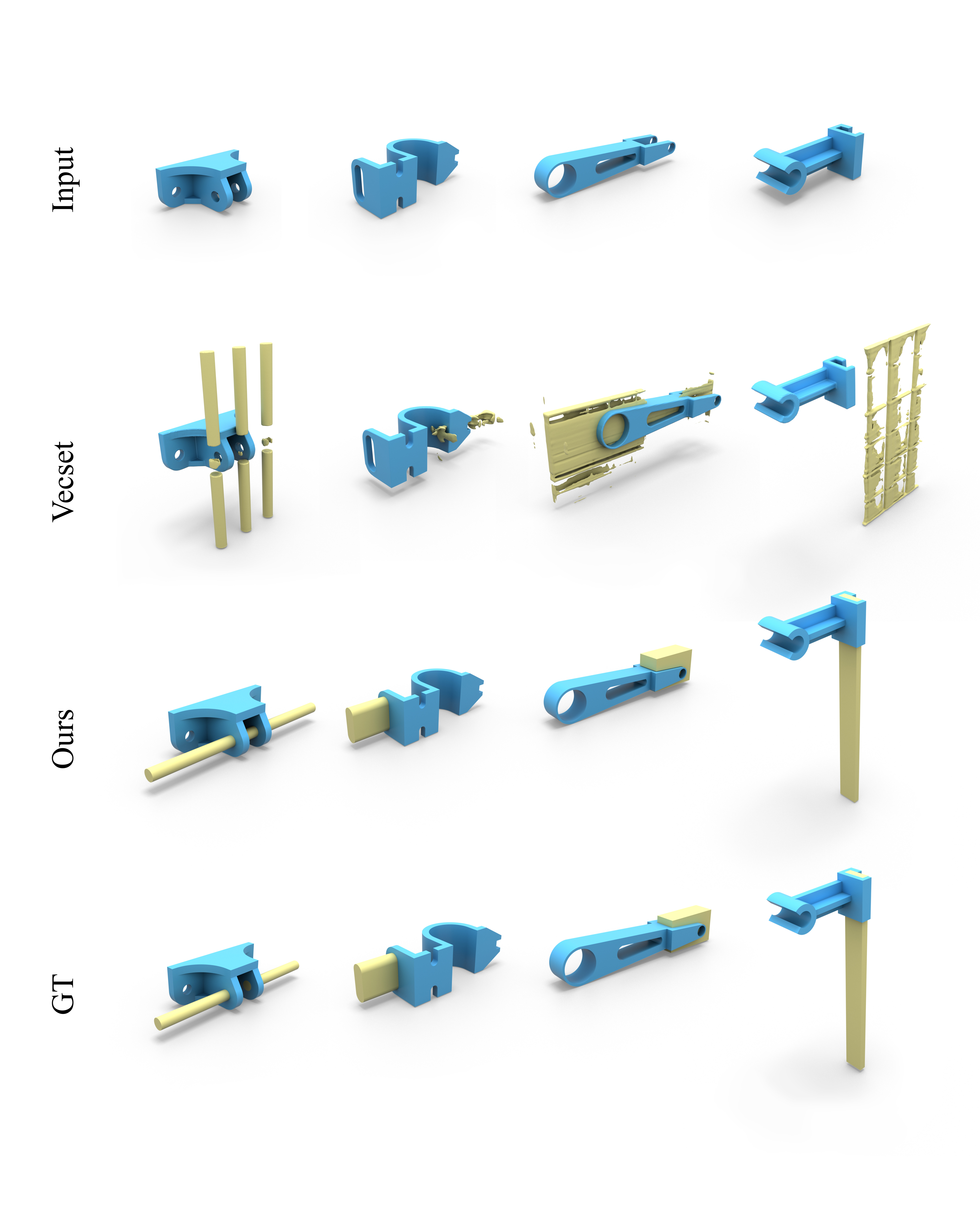}
  \caption{Visual comparison on MoldCAD dataset.}
\label{fig:results_comparision}
\end{figure}
\subsection{Data Sourcing and Filtering}
We initialized our candidate pool using the ABC ~\cite{Koch_2019_CVPR} and BRepNet ~\cite{lambourne2021brepnet} datasets, which provide a rich variety of industrial CAD models. However, raw data from these repositories often contains artifacts that cannot be manufactured by injection molding (e.g., non-demoldable geometries, multi-body parts) or simple geometries that do not require complex molding strategies. We first filtered single-body parts that satisfy basic injection molding prerequisites, such as water-tightness and appropriate wall thickness, and then explicitly retained and prioritized geometries with prominent undercuts and side holes. These features necessitate auxiliary molding components, rather than simple two-piece molds, thereby ensuring that the dataset captures the inherent challenges of actual manufacturing.
\input{tables/gen_fidelity_combined}

\subsection{Professional Mold Design and Annotation}
For each curated part, we commission professional mold engineers to design a complete injection-mold assembly following standard industrial practice. Figure~\ref{fig:mold_design_pipeline} summarizes the typical workflow on a representative part with such features. Engineers first select a draft direction and analyze demolding capability along that direction to identify areas that might be mechanically locked during upper and lower mold separation. In practice, these problematic regions often appear as concave or side features relative to the mold opening motion. To resolve these geometric constraints, engineers design auxiliary components by determining feasible translation vectors and segmenting the geometry to ensure collision-free movement. Finally, the parting surface is constructed to split the mold into two halves. The resulting assemblies reflect real design decisions made in production settings and serve as precise ground truth for learning-based generation.

Each mold assembly is annotated at the instance level, providing distinct labels for the upper mold, lower mold, parting surfaces, and all auxiliary components. To handle the combinatorial complexity introduced by auxiliary mechanisms, MoldCAD explicitly preserves these parts as discrete entities rather than merging them into a single geometry. It enables models to learn discrete part generation rather than mere surface reconstruction and supports more granular evaluation metrics.

%% file: tables/gen_fidelity_combined.tex
\begin{table}[t]  
	\centering  
	\scriptsize
    \setlength{\tabcolsep}{4pt}  
    \begin{minipage}[t]{0.49\textwidth}
        \centering
        \begin{tabular}{c|ccc}  
            \toprule
            Method & $\underset{\%\uparrow}{\text{COV}}$ & $\underset{\downarrow}{\text{MMD}}$ & $\underset{\downarrow}{\text{JSD}}$ \\
            \midrule  
            VecSet-based & 21.24 & 0.02900  & 0.105 \\  
            Ours-only first stage & 55.68 & \textbf{0.0140} & 0.054\\ 
            AIMold & \textbf{57.29} & 0.0147 & \textbf{0.050}  \\  
            \bottomrule
        \end{tabular}  
        \caption{Auxiliary keys generation fidelity of different representations method. }
        \label{tab:aux generation_fidelity}  
    \end{minipage}%
    \hspace{5pt}%
    \begin{minipage}[t]{0.49\textwidth}
        \centering
        \begin{tabular}{c|ccc}  
            \toprule
            Method & $\underset{\%\uparrow}{\text{COV}}$ & $\underset{\downarrow}{\text{MMD}}$ & $\underset{\downarrow}{\text{JSD}}$ \\
            \midrule  
            VecSet-based & 9.21 & 0.0252 & 0.277 \\  
            Ours-only first stage &46.55  & 0.0096 & 0.073\\ 
            AIMold & \textbf{49.01} & \textbf{0.0091} & \textbf{0.061}  \\  
            \bottomrule
        \end{tabular}  
        \caption{Upper and lower mold generation fidelity of different representations method. }
        \label{tab:uplow generation_fidelity}  
    \end{minipage}
\end{table}

%% file: sec/5_experiment.tex
\section{EXPERIMENTS}
\subsection{Implementation}
For the two-stage training, we use our carefully constructed MoldCAD dataset. In the first-stage training, we adopt the AdamW optimizer ~\cite{loshchilov2017decoupled} with a learning rate of $1\times10^{-4}$ and follow ~\cite{trellis} setup in second-stage geometry flow model training. For structure latents encoding, we render 12 images per CAD model. All experiments are conducted on NVIDIA A100 GPUs. The dataset is split into training and test sets with a ratio of 9:1, where the test set accounts for 10\% of the total data and is used for evaluation.
\subsection{Comparison}

\subsubsection{Quantitative Evaluation.}
Tables~\ref{tab:aux generation_fidelity} and~\ref{tab:uplow generation_fidelity} report the averaged generation quality scores for auxiliary components, upper and lower molds on our MoldCAD dataset. We primarily compare representative implicit methods operating in the latent space, which employ VecSet-based representation architectures like~\cite{cube3d,zhao2025hunyuan3d}. These approaches have demonstrated strong performance on large-scale 3D datasets. However, in the context of mold design tasks, the lack of sufficiently large and diverse datasets makes it difficult for latent-space methods to achieve proper alignment, as evidenced by the results in Figure~\ref{fig:results_comparision} and Table~\ref{tab:aux generation_fidelity}. AIMold consistently outperforms both baselines across almost all metrics, achieving better COV, MMD, and JSD scores (details in Appendix A). These results indicate improved generation quality and a closer match to the ground-truth data distribution. Furthermore, the second-stage voxel refinement leads to more accurate quantitative metrics and produces CAD models with more evident geometric features.

For orientation estimation, directly regressing a low-dimensional orientation vector from a high-dimensional 3D representation is highly challenging. As shown in Table~\ref{tab:orientation acc}, a naive regression-based approach fails to produce reliable results and exhibits very poor performance. 
In contrast, our plane-detection-based method significantly reduces the solution space by discretizing the continuous orientation domain into a small set of candidate orientations. This strategy effectively addresses the orientation estimation problem and provides a robust and accurate orientation initialization for the subsequent generation pipeline.

\input{tables/orien_main}
\subsection{Ablations}
\input{tables/axu_gen_ablate}
\subsubsection{Two-stage Generation} To demonstrate the effectiveness of our generation approach, we conduct an ablation study to examine whether both stages of the pipeline are necessary. Specifically, we remove the second stage and retain only the first stage. This modification not only eliminates the ability to convert voxel representations into mesh, but also leads to a noticeable degradation in quantitative performance as shown in Table~\ref{tab:aux generation ablate}. In addition, we investigate a weakened variant of the second stage by reducing the number of DINO feature views used for structural latent generation. This ablation further degrades geometric accuracy, indicating that sufficient multi-view features are critical for accurate structure modeling. The "Less Render View" setting uses only four views to construct the structural latent representations.

\subsubsection{3D Encoder Pretraining} To study the impact of large-scale 3D data pretraining for our task, we employ a 3D encoder~\cite{trellis} pretrained on a large-scale dataset such as~\cite{deitke2023objaverse}. As shown in Table~\ref{tab:aux generation ablate}, without pretraining, the performance clearly degrades, indicating that current large-scale 3D data provides a valuable prior for our mold design task. 

\subsection{Failure Cases}
AIMold suffers from two common categories of failure cases: 1) Thin or small structures, which are challenging due to resolution limitations. Future improvements in resolution efficiency could help mitigate this issue. 2) Missing faces, resulting in non-watertight solids, a failure that is anticipated, as AIMold cannot guarantee watertight outputs. Optimization with representations that guarantee watertightness could serve as a potential solution.

%% file: tables/orien_main.tex
\begin{table}[t]  
	\centering  
	\scriptsize
    \vspace{-8pt}
    \setlength{\tabcolsep}{4pt}  
	\begin{tabular}{c|ccc}  
        \toprule
        Method & $\underset{\%\uparrow}{\text{Accuracy}}$ & $\underset{\%\uparrow}{\text{Recall}}$ & $\underset{\uparrow}{\text{F1-Score}}$ \\
        \midrule   
        Regressive-based & 12.66 & 12.79 & 0.2267 \\ 
        \textbf{Ours} & 91.17 & 89.65 & 0.9129 \\  
        \bottomrule
	\end{tabular}  
\caption{Model orientation estimation accuracy of different method. }
    \vspace{-8pt}
	\label{tab:orientation acc}  
\end{table}

%% file: tables/axu_gen_ablate.tex
\begin{table}[t]  
	\centering  
	\scriptsize
    \vspace{-8pt}
    \setlength{\tabcolsep}{4pt}  
	\begin{tabular}{c|ccc}  
        \toprule
        Method & $\underset{\%\uparrow}{\text{COV}}$ & $\underset{\downarrow}{\text{MMD}}$ & $\underset{\downarrow}{\text{JSD}}$ \\
		 \midrule  
        From Scratch & 52.82 & 0.0173  & 0.0682 \\  
        Less Render View & 55.44 & 0.0146  & 0.050 \\
		Ours-only first stage & 55.68 & \textbf{0.0140} & 0.054\\ 
		\textbf{Ours} & \textbf{57.29} & 0.0147 & \textbf{0.050}  \\  
		\bottomrule
	\end{tabular}  
    \caption{Auxiliary keys generation fidelity ablation study. }
    \vspace{-8pt}
	\label{tab:aux generation ablate}  
\end{table}

%% file: sec/6_limitation_futurework.tex
\section{LIMITATIONS AND FUTURE WORK}
Although our MoldCAD dataset for the mold design task has already yielded promising results at its current scale, the scale of MoldCAD still exhibits a significant gap when compared to primarily single-body CAD datasets~\cite{wu2021deepcad,Koch_2019_CVPR} or open-vocabulary 3D datasets~\cite{deitke2023objaverse}. The scale of our curated dataset, MoldCAD, can be further improved with additional effort. We observe the trade-off in the number of render views used when transferring the mesh to the structured latent space, as shown in Table \ref{tab:aux generation ablate}, which limits the capture of occlusion-related structural information. Exploring more efficient structure encoding methods or improved representations may help mitigate this trade-off.

%% file: sec/7_conclusion.tex
\section{CONCLUSION}
We introduce AIMold, a conditional generation model for complex mold design. We present our newly collected MoldCAD dataset and conduct extensive experiments to validate the effectiveness of our method. By decomposing the complex design task into structured generation steps, we address the challenges inherent in mold design. We believe this approach demonstrates a promising pathway for predicting demolding orientations, identifying auxiliary components, and constructing parting surfaces, ultimately enabling the generation of a complete, manufacturing-ready mold assembly.

%% file: sec/8_ack.tex
\section{Acknowledgments}
The work was supported in part by Guangdong S\&T Programme with Grant No. 2024B0101030002, the Basic Research Project No. HZQB-KCZYZ-2021067 of Hetao Shenzhen-HK S\&T Cooperation Zone, the Shenzhen Outstanding Talents Training Fund 202002, the NSFC with Grant No. 62293482, the Guangdong Provincial Key Laboratory of Future Networks of Intelligence (Grant No. 2022B1212010001), the Shenzhen Key Laboratory of Big Data and Artificial Intelligence (Grant No. SYSPG20241211173853027) , the Guangdong Province Radio Science Data Center with grant No. 2025B1212070001, the National Key R\&D Program of China with grant No. 2018YFB1800800.

%% file: sec/X_suppl.tex
\newpage
\appendix
\label{sec:append_a}
\section{DETAILS OF EVALUATION}

For auxiliary components, upper and lower molds generation, we adopt distribution-based metrics~\cite{xu2024brepgen} to quantitatively evaluate generation quality. For each 3D model, we uniformly sample 2,000 points from the surface and compute the following metrics: \textit{Coverage} (COV), \textit{Minimum Matching Distance} (MMD), and \textit{Jensen–Shannon Divergence} (JSD). COV measures the percentage of reference shapes that are matched by at least one generated shape, where each generated shape is assigned to its nearest neighbor in the reference set based on the Chamfer Distance (CD). MMD is defined as the average CD between each reference shape in the reference set and its nearest neighbor in the generated set. JSD quantifies the distributional discrepancy between the reference and generated data by voxelizing the corresponding point clouds into $28^3$ discrete grids.

\section{REAL-WORLD VALIDATION}

\begin{figure}[h!]
  \centering
  \scalebox{0.95}{%
    \begin{minipage}{0.45\columnwidth}
      \centering
      \includegraphics[width=\linewidth]{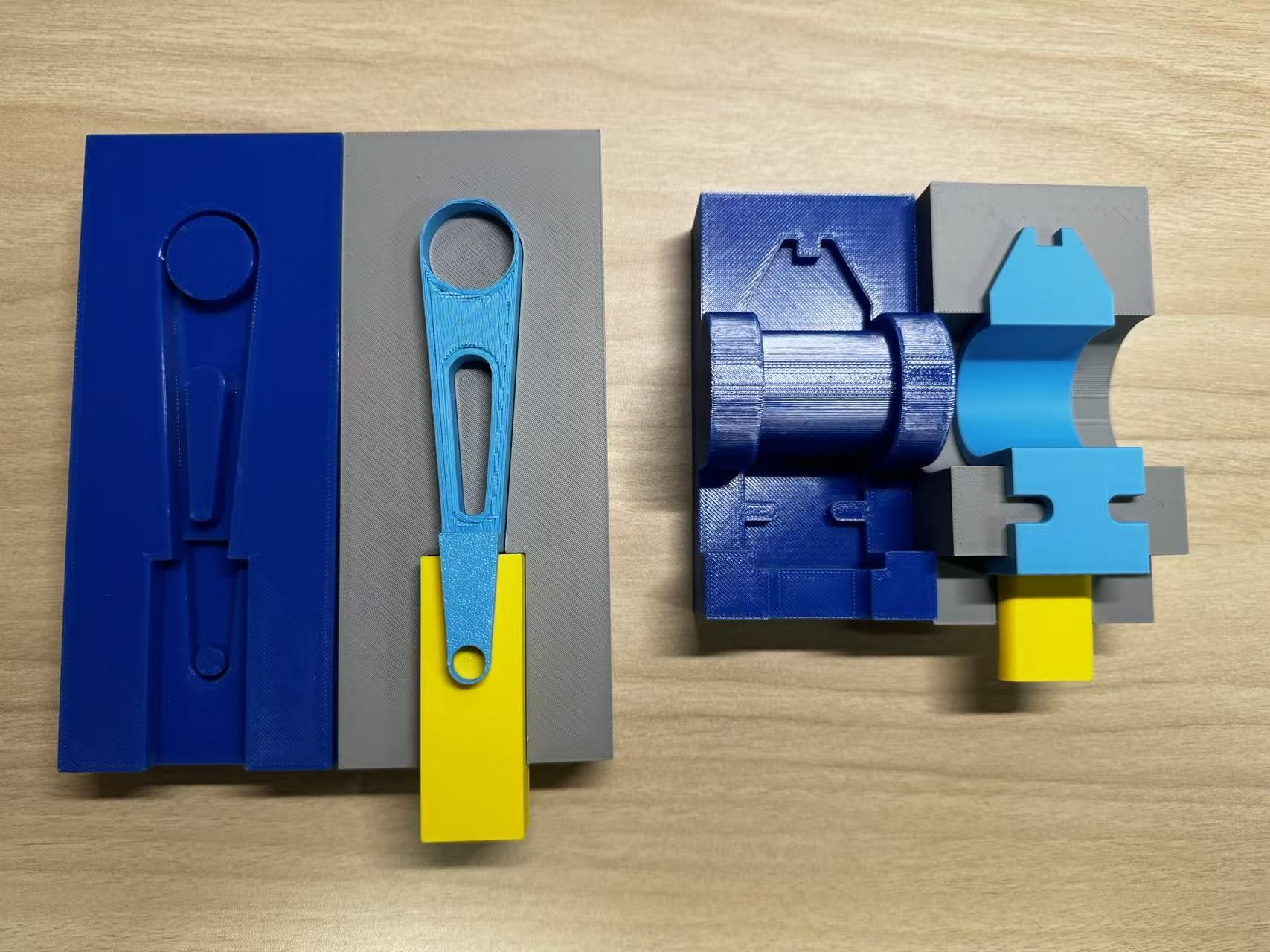}
    \end{minipage}%
    \hspace{0.5cm}%
    \begin{minipage}{0.45\columnwidth}
      \centering
      \includegraphics[width=\linewidth]{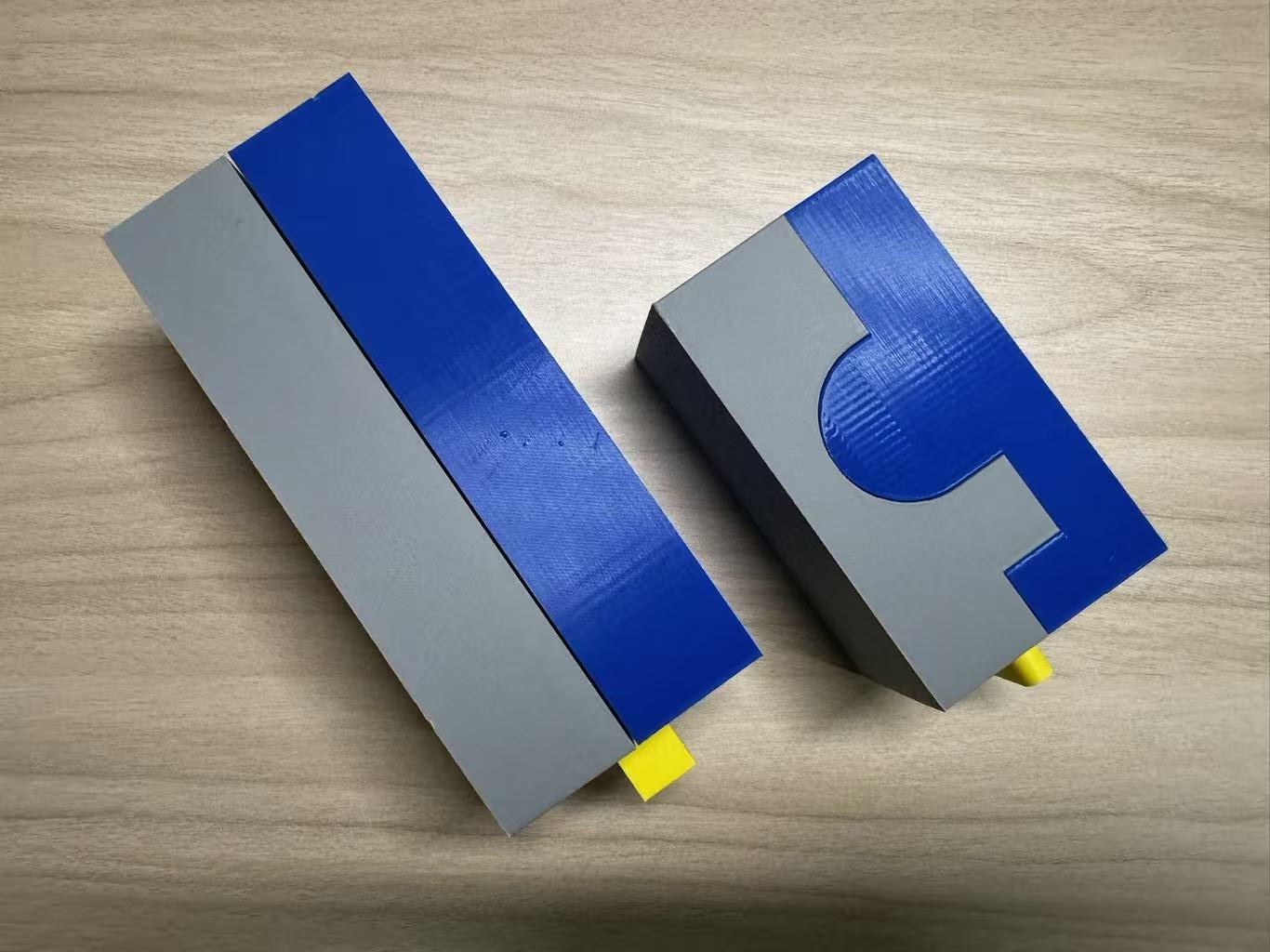}
    \end{minipage}%
  }
  \footnotesize
  \caption{Real-world validation.} 
  \label{fig:mold}
\end{figure}

To provide an additional physical check, a 3D printing prototype is also presented in Fig.~\ref{fig:mold} to physically validate assembly and demolding behavior. We use 3D printing because actual fabrication at an industrial mold factory requires weeks of lead time, making it infeasible during the short rebuttal period.

\newpage
\section{NECESSITY OF UPPER/LOWER MOLD LEARNING}

\begin{figure}[h!]
\begin{minipage}{0.95\columnwidth}
  \centering
\includegraphics[width=0.85\linewidth]{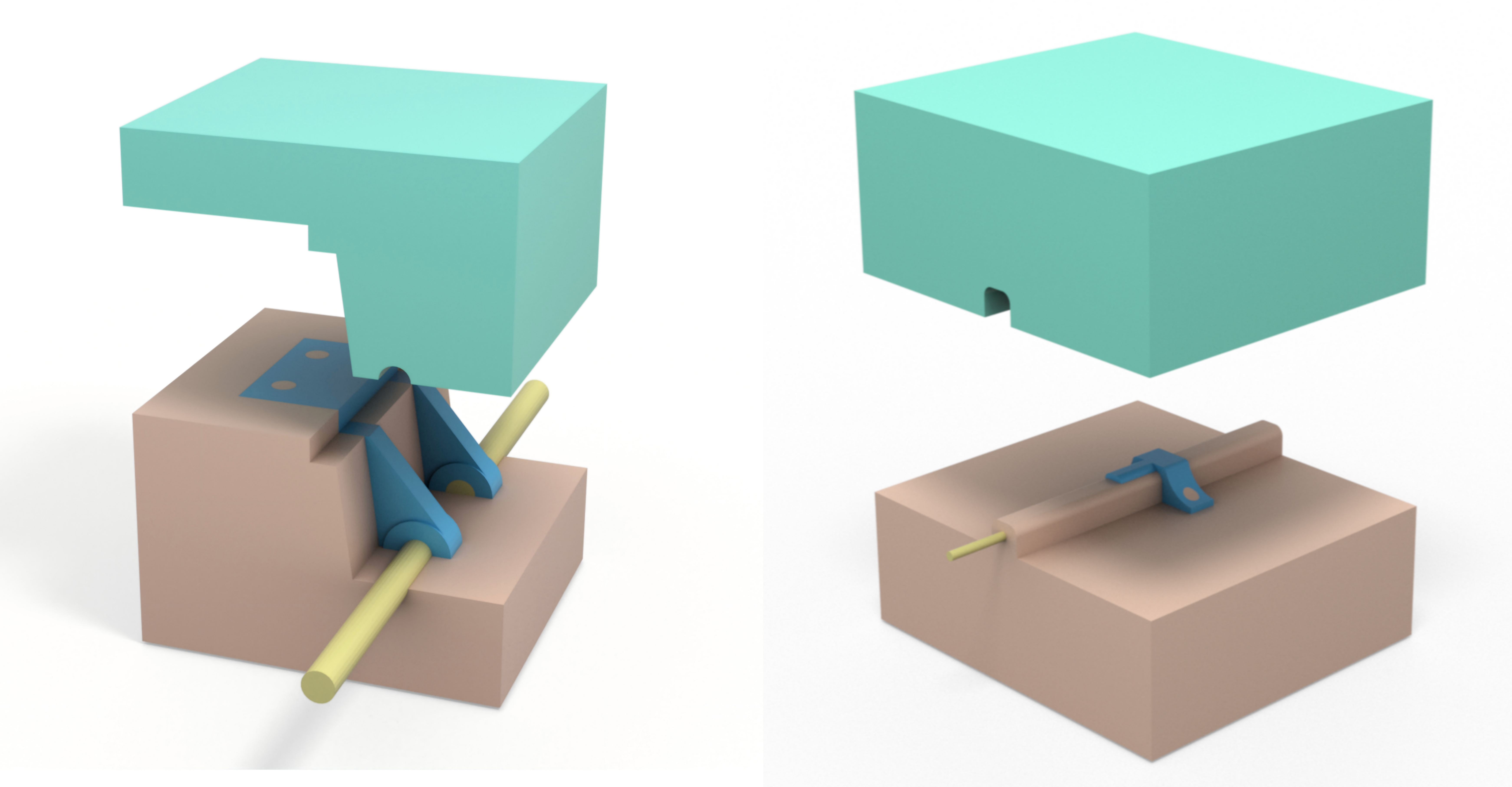}
  \footnotesize
  \caption{Complex parting surface examples.}
  \label{fig:example_pl}
\end{minipage}
\end{figure}

In actual mold design, the parting surface is not given and is sometimes not a simple plane, as shown in Fig.~\ref{fig:example_pl}. For complex models, using a planar parting surface can cut through important surface features, create unnatural mold boundaries, or produce thin/weak mold regions near holes and side structures, which affects appearance quality, structural robustness, and geometric accuracy. Therefore, learning-based method is necessary for handling complex, non-planar parting surfaces.

